\documentclass{article}
\usepackage{spconf,amsmath,graphicx}
\usepackage{amsfonts}
\usepackage{arydshln}
\usepackage{bbm}
\usepackage{color}
\usepackage{epstopdf}
\usepackage{enumitem}
\usepackage{tikz}

\tikzset{>=latex}

\newcommand{\bphi}{\boldsymbol{\phi}}

\newcommand{\btheta}{\boldsymbol{\theta}}

\newcommand{\argmax}{\mathop{\text{argmax}}}
\newcommand{\cost}{\mathop{\text{cost}}}
\newcommand{\head}{\text{head}}
\newcommand{\tail}{\text{tail}}

\newcommand{\term}[1]{\textbf{#1}}

\newcommand{\HTcomment}[1]{}
\newcommand{\WWcomment}[1]{}
\newcommand{\KGcomment}[1]{}
\newcommand{\KLcomment}[1]{}

\title{DISCRIMINATIVE SEGMENTAL CASCADES\\FOR FEATURE-RICH PHONE RECOGNITION}
\name{Hao Tang, Weiran Wang, Kevin Gimpel, Karen Livescu}
\address{Toyota Technological Institute at Chicago\\
    {\small\texttt{\{haotang,weiranwang,kgimpel,klivescu\}@ttic.edu}}}

\begin{document}
\maketitle

\begin{abstract} 

%\KLcomment{Did some reworking on terminology.  To make things match throughout paper, SCRF $\longrightarrow$ ``discriminative segmental model'' (or just ``segmental model'') for the general model class, ``structured cascade'' for our models, ``inference'' $\longrightarrow$ ``decoding''?}
Discriminative segmental models, such as segmental conditional random fields (SCRFs) and segmental structured support vector machines (SSVMs), have had success in speech recognition via both lattice rescoring and first-pass decoding.  However, 
% SCRFs 
such models suffer from slow 
%inference, 
decoding, hampering the use of computationally
expensive features, such as 
% segmental 
segment neural networks or other high-order features.
%Meanwhile, several approximate inference algorithms have been proposed
%for general graphical models, e.g.,
%beam search, Lagrangian relaxation, and structured prediction cascades (SPC).  
A typical solution 
%approach for computationally demanding decoding 
is to use approximate decoding, either by beam pruning in a single pass or by beam pruning to generate a lattice followed by a second pass.
%, can be considered for SCRFs.
In this work, we study 
%them in the context of training SCRFs with a hinge loss.
discriminative segmental models trained with a hinge loss (i.e., segmental structured SVMs).
We show that beam search is not suitable for learning rescoring models in this approach, though it 
%; beam search however 
gives good approximate decoding performance when the model is already well-trained.
Instead, we consider an approach inspired by structured prediction cascades, which use max-marginal pruning to generate lattices.  
We obtain a high-accuracy phonetic recognition system with several expensive feature types: a segment neural network, a second-order language model, and second-order phone boundary features.

\end{abstract}

\begin{keywords}
segmental conditional random field, structured prediction cascades, phone recognition, segment neural network, beam search
\end{keywords}

\vspace{-0.1in}
\section{INTRODUCTION}
\label{sec:intro}

Segmental models have been considered for speech recognition as an alternative to frame-based models such as hidden Markov models (HMMs), in order to address the shortcomings of the frame-level Markov assumption and introduce expressive segment-level features.  Segmental models include \term{segmental conditional random fields} (SCRFs) \cite{ZweigNguyen2009}, or semi-Markov conditional random fields~\cite{SarawagiCohen2004}; \term{segmental structured support vector machines} (SSVMs) \cite{ZhangGales2013}; and generative segmental models~\cite{OstendorfEtAl1996,Glass2003}.  Previous work comparing segmental model training algorithms has shown some benefits of discriminative segmental models trained with hinge loss (SSVM-type learning)~\cite{TangEtAl2014}, and we consider this type of model here.

%Various types of expressive segment-level features have been considered, mainly at the word level~\cite{ZweigEtAl2011} \KLcomment{cite some others?} but also at the phone level~\cite{Zweig2012,HeFoslerLussier2012,TangEtal2014}.

Discriminative segmental models have allowed the exploration of complex features, both at the word level~\cite{ZweigEtAl2011} and at the phone level~ \cite{Zweig2012, HeFoslerLussier2012, TangEtAl2014}.
%Several complex word-level features have been explored in the past \cite{ZweigEtAl2011}.
%However, few at the phone level have been explored in the context of segmental models
%except in \cite{Zweig2012, HeFoslerLussier2012, TangEtAl2014}.
%An intuitive approach is to use the output of a phone segment classifier as features in an segmental model.
%Phone segment classification was once a popular task \cite{AndenMallat2014, ChangGlass2007, ClarksonMoreno1999,
%Halberstadt1998, LeungEtAl1992, ZahorianEtAl1997},
%but its incompatibility with frame-based models has led to
%a loss of interest.  In this paper, we revisit the question of whether segment classifiers are able
%to improve the downstream recognition task.  In particular, we investigate and revive the use of
%\term{segment neural networks} \cite{LeungEtAl1992, AustinEtAl1992, ZahorianEtAl1997}. 
%\KGcomment{I suggest we might want to move this paragraph and/or rewrite it once we know our full experimental story.} \KLcomment{yes}
These powerful segmental features are a double-edged sword---on
the one hand, the model becomes more expressive; on the other,
it is computationally challenging to decode with and train such models. 
%\KLcomment{trying to use more speech-y terminology like ``decoding'' instead of ``inference''}
%infer and to train. 
For this reason, SCRFs~\cite{ZweigNguyen2010} and SSVMs~\cite{ZhangGales2013} were initially applied to speech recognition in a multi-pass approach, where the segmental model considers only a subset of the hypothesis space contained in lattices generated by HMMs.
Much effort has been devoted to removing the dependency on
HMMs and instead developing \term{first-pass segmental models}
\cite{AbdelHamidEtAl2013, HeFoslerLussier2012, HeFoslerLussier2015}.
However, working with the entire hypothesis space 
imposes an even larger burden on inference, especially
when the features are computationally intensive or of high order.

If we wish to consider the entire search space in decoding, we can only afford features
of low order or of specific types as in \cite{HeFoslerLussier2012}.
An alternative approach to the problem is to use approximate
decoding.  There are two widely used approximate decoding
algorithms: beam search and multi-pass decoding.
In the intuitive and popular beam search, 
the idea is to prune as we search along the graph representing the search space.
It has been used for decoding in almost all HMM systems,
and for generating lattices as well.
Though popular, it offers no guarantees
about its approximation.
In the category of multi-pass decoding, lattice and $n$-best list rescoring~\cite{OstendorfEtAl1991} are commonly used alternatives. \KLcomment{note to check later:  is \cite{OstendorfEtAl1991} lattice rescoring?}
%If decoding with
%complex features for the entire hypothesis space
%is too expensive, a subset of the hypothesis space
%, or lattice, 
%can be generated
%with a simpler model and rescored with the desired complex features.

We focus on a particular type of multi-pass approach based on 
structured prediction cascades \cite{WeissEtAl2012}, which
we term \term{discriminative segmental cascades}.  
A cascade is a general approach
for decoding and training complex structured models,
using a multi-pass sequence of models with increasing order of features, while
pruning the hypothesis space by a multiplicative factor to counteract
the growth in feature computation.
\KLcomment{want to check if the last statement is precise.  In our case it is not just an issue of exponential growth but also expensive features in general.}
\HTcomment{better?}
In this approach, the hypothesis space in each pass is pruned with \term{max-marginals},
which offers the guarantee that all paths with scores higher than the pruning
threshold are kept.  

Applying the discriminative segmental cascade approach to speaker-independent phonetic recognition on the TIMIT data set, we obtain a first-pass phone error rate of 21.4\% with a unigram language model, and a two-stage cascade error rate of 19.9\%, which includes a bigram language model, a segment neural network classifier, and second-order phone boundary features.  This is to our knowledge the best result to date with a segmental model.  In the following sections we define the discriminative segmental models we consider, describe how we represent a cascade of hypothesis spaces with a finite-state composition-like operation, present discriminative segmental cascades for decoding and training with max-marginal pruning, and discuss our experiments.
%\KGcomment{No preview on the experimental results? I'd prefer a sentence or two previewing our most interesting results instead of an outline of the rest of the paper, since I think our results are interesting and I tend to skip ahead whenever I see any outline-like text :)}

\section{DISCRIMINATIVE SEGMENTAL MODELS}
\label{sec:scrf}

A \term{linear segmental model} for input space $\mathcal{X}$
and hypothesis space $\mathcal{Y}$ is defined 
%abstractly 
formally 
as a pair $(\btheta, \bphi)$, where
$\btheta \in \mathbb{R}^d$ is the parameter vector and 
$\bphi: \mathcal{X} \times \mathcal{Y} \to \mathbb{R}^d$ is
the feature vector.  For an input 
%utterance 
$x \in \mathcal{X}$,
each hypothesis $y \in \mathcal{Y}$ is associated
with a score $\btheta^\top \bphi(x, y)$, and the goal
of decoding is to find the hypothesis that maximizes
the score,
% or equivalently solve
\begin{equation} \label{eq:inf}
\argmax_{y \in \mathcal{Y}} \,\,\btheta^\top \bphi(x, y).
\end{equation}
\KLcomment{hmm, so far this is just a general linear model... nothing segmental or even structured here}
For speech recognition, we formally define the hypothesis space $\mathcal{Y}$ in terms of finite-state transducers (FST). 
Let $\Sigma$ be the label set (e.g., the phone set in phone recognition),
and $\overline{\Sigma} = \Sigma \cup \{\epsilon\}$,
where $\epsilon$ is the empty label.
Define a \term{decoding graph} as a standard FST
$G = (V, E, I, F, w, i, o)$, where $V$ is the set
of vertices, $E \subseteq V \times V$ is the set
of edges, $I \subseteq V$ is the set of initial vertices,
$F \subseteq V$ is the set of final vertices,
$w: E \to \mathbb{R}$ is a function that associates a weight to an
edge, $i: E \to \overline{\Sigma}$ is a function that associates an input
label to an edge,
and $o: E \to \overline{\Sigma}$ is a function that associates an output
label to an edge.
In addition to the standard definition of FSTs, we equip $G$ with
a function $t: V \to \mathbb{R}$ that maps a vertex to a time stamp.
For any edge $(u, v) \in E$, let $\tail((u, v)) = u$, and $\head((u, v)) = v$.
For convenience, we will use subscripts to denote components
of a particular FST, e.g., $E_G$ is the edge set of $G$.

For an input utterance, let $x$ be the sequence of acoustic feature
vectors.
We construct a decoding graph $G$ from $x$, then define our hypothesis space $\mathcal{Y} \subseteq 2^E$ to be the subset of
paths
%a subset of the power set of $E$,
that start at an initial vertex in $I$ and end at a final vertex in $F$. 
%, is our hypothesis space.
A path $y \in \mathcal{Y}$ of length $m$ is a sequence of unique edges
$\{e_1, \dots, e_m\}$,
satisfying $\head(e_i) = \tail(e_{i+1})$ for $i \in [m]$.
Given a model $(\btheta, \bphi)$, for each edge $e \in E$,
the weight $w(e)$ is defined as $\btheta^\top \bphi(x, e)$.
For convenience, for a path $y \in \mathcal{Y}$, we overload $\bphi$ and $w$ and define
$\bphi(x, y) = \sum_{e \in y} \bphi(x, e)$
and $w(y) = \btheta^\top \bphi(x, y) = \sum_{e \in y} w(e)$,
where we treat a path $y$ as a set of (unique) edges $e$.

\begin{figure*}
\begin{center}
\begin{tikzpicture}[ver/.style={draw,circle}]
\node [ver] (x1) at (0, 0) {};
\node [ver] (x2) at (1, 0) {};
\node [ver] (x3) at (2, 0) {};
\node [ver] (x4) at (3, 0) {};
\node [ver] (x5) at (4, 0) {};

\draw[->,black!30] (x1) to [out=10,in=170] (x2);
\draw[->,black!30] (x1) to [] (x2);
\draw[->,black!30] (x1) to [out=-10,in=190] (x2);

\draw[->,black!30] (x2) to [out=10,in=170] (x3);
\draw[->,black!30] (x2) to [] (x3);
\draw[->,black!30] (x2) to [out=-10,in=190] (x3);

\draw[->,black!30] (x3) to [out=10,in=170] (x4);
\draw[->,black!30] (x3) to [] (x4);
\draw[->,black!30] (x3) to [out=-10,in=190] (x4);

\draw[->,black!30] (x4) to [out=10,in=170] (x5);
\draw[->,black!30] (x4) to [] (x5);
\draw[->,black!30] (x4) to [out=-10,in=190] (x5);

\draw[->,black!30] (x1) to [out=30,in=150] (x3);
\draw[->,black!30] (x1) to [out=25,in=155] (x3);
\draw[->,black!30] (x1) to [out=20,in=160] (x3);

\draw[->,black!30] (x2) to [out=30,in=150] (x4);
\draw[->,black!30] (x2) to [out=25,in=155] (x4);
\draw[->,black!30] (x2) to [out=20,in=160] (x4);

\draw[->,black!30] (x3) to [out=30,in=150] (x5);
\draw[->,black!30] (x3) to [out=25,in=155] (x5);
\draw[->,black!30] (x3) to [out=20,in=160] (x5);

\draw[->,black!30] (x1) to [out=45,in=135] (x4);
\draw[->,black!30] (x1) to [out=40,in=140] (x4);
\draw[->,black!30] (x1) to [out=35,in=145] (x4);

\draw[->,black!30] (x2) to [out=45,in=135] (x5);
\draw[->,black!30] (x2) to [out=40,in=140] (x5);
\draw[->,black!30] (x2) to [out=35,in=145] (x5);

\node [ver] (y1) at (5, 0) {};
\node [ver] (y2) at (6, 0) {};
\node [ver] (y3) at (7, 0) {};
\node [ver] (y4) at (8, 0) {};
\node [ver] (y5) at (9, 0) {};

\draw[->,black!30] (y1) to [out=20,in=160] node [black] {a} (y2);
\draw[->,black!30] (y1) to [out=-20,in=200] node [black] {b} (y2);

\draw[->,black!30] (y4) to [] node [black] {c} (y5);
\draw[->,black!30] (y2) to [out=35,in=145] node [black] {b} (y4);
\draw[->,black!30] (y2) to node [black] {a} (y3);
\draw[->,black!30] (y3) to node [black] {a} (y4);

\node [ver] (a1) at (10.5, 0) {};
\node [ver] (a2) at (11.5, 0) {};
\node [ver] (a3) at (10.5, 1) {};
\node [ver] (a4) at (11.5, 1) {};

\draw[->,black!30] (a3) to (a1);
\draw[->,black!30] (a3) to (a2);
\draw[->,black!30] (a3) to (a4);

\draw[->,black!30] (a1) to [out=55,in=-145] (a4);
\draw[->,black!30] (a4) to [out=-125,in=35] (a1);

\draw[->,black!30] (a1) to [out=10,in=170] (a2);
\draw[->,black!30] (a2) to [out=190,in=-10] (a1);

\draw[->,black!30] (a2) to [out=100,in=-100] (a4);
\draw[->,black!30] (a4) to [out=-80,in=80] (a2);

\draw[->,black!30] (a2) to [loop right] (a2);
\draw[->,black!30] (a1) to [loop left] (a1);
\draw[->,black!30] (a4) to [loop right] (a4);

\node [ver] (z1) at (13, 0) {};
\node [ver] (z21) at (14, 0) {};
\node [ver] (z22) at (14, 1) {};
\node [ver] (z3) at (15, 0) {};
\node [ver] (z41) at (16, 0) {};
\node [ver] (z42) at (16, 1) {};
\node [ver] (z5) at (17, 0) {};

\draw[->,black!30] (z1) to node [black] {$\epsilon$\_a} (z21);
\draw[->,black!30] (z1) to node [black] {$\epsilon$\_b} (z22);

\draw[->,black!30] (z21) to node [black] {a\_a} (z3);
\draw[->,black!30] (z21) to node [black] {a\_b} (z42);
\draw[->,black!30] (z22) to node [black] {b\_a} (z3);
\draw[->,black!30] (z22) to node [black] {b\_b} (z42);

\draw[->,black!30] (z3) to node [black] {a\_a} (z41);
\draw[->,black!30] (z41) to node [black] {a\_c} (z5);
\draw[->,black!30] (z42) to node [black] {b\_c} (z5);

\node at (2, -0.6) {$H_1$};
\node at (7, -0.6) {$H_2$};
\node at (11, -0.6) {$L_2$};
\node at (15, -0.6) {$H_2 \circ_\sigma L_2$};
\end{tikzpicture}
\caption{\label{fig:scrf} \emph{From left to right}: An example of the full hypothesis space
$H_1$ with four frames (five vertices) and three unique labels \{a, b, c\}
(three edges between every pair of vertices) with segment length up to three
frames (actual labels omitted for clarity); $H_2$, a pruned $H_1$; a graph structure corresponding to a bigram language model $L_2$ over three labels; and $H_2$ $\sigma$-composed with $L_2$, where $s_1$\_$s_2$
denotes the bigram $s_1s_2$. \KLcomment{why not use $s_1s_2$?}}
\end{center}
\vspace{-0.2in}
\end{figure*}
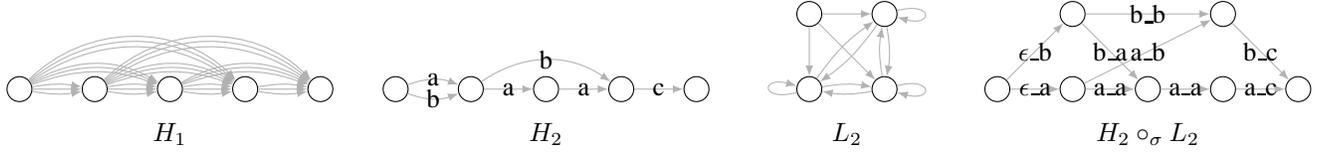

If the decoding graph is the full hypothesis space with all possible
segmentations and all possible labels, for example the graph
on the left in Figure~\ref{fig:scrf},
then the model is a \term{first-pass segmental model}.  Otherwise,
it is a lattice rescoring model.
By the above definitions, inference (decoding) in the model \eqref{eq:inf}
can be solved with a standard shortest-path algorithm.

The model parameters $\btheta$ can be learned by minimizing
the sum of loss functions on samples $(x,y)$ in a training set.
In general, the model can be trained with different losses.
The model is an SCRF if we train it with log loss $-\log p(y | x)$
where $p(y|x) \propto \exp(\btheta^\top \bphi(x, y))$.
It is a segmental structured SVM if we use the structured hinge loss:
\begin{equation}
\ell_{\text{hinge}}(\btheta) = \max_{y' \in \mathcal{Y}}\bigg[\cost(y, y')
    - \btheta^\top \bphi(x, y)
    + \btheta^\top \bphi(x, y')\bigg],
\end{equation}
where $\cost: \mathcal{Y} \times \mathcal{Y} \to [0, \infty)$
measures the badness of a hypothesis path $y'$
compared with the ground truth $y$.

The loss can be optimized with first-order methods,
such as stochastic gradient descent (SGD).  The gradient (or subgradient, in this case) computation typically
involves a forward-backward-like algorithm.  For example,
the subgradient of the hinge loss is
\begin{equation}
\nabla_{\btheta} \ell_{\text{hinge}}(\btheta) = -\bphi(x, y) + \bphi(x, \tilde{y}), 
\end{equation}
where computing the \term{cost augmented path}
\begin{equation}
\tilde{y} = \argmax_{y' \in \mathcal{Y}}\; \cost(y, y') + \btheta^\top \bphi(x, y'),
\end{equation}
requires a forward pass over the graph.
Compared to computing the gradient
of other losses, which requires more forward passes and backward passes,
hinge loss has computational advantages,
and has been shown to perform well \cite{TangEtAl2014},
so we will use hinge loss for the rest of the paper.

\section{HIGH-ORDER FEATURES \\ AND STRUCTURED COMPOSITION}
\label{sec:feat}

The order of a feature is defined as the number of 
labels on which it depends.
A feature is said to be a \term{first-order
feature} if it depends on a single label,
a \term{second-order feature} if it depends
on a pair of labels, and so on.
Features with no label dependency are called \term{zeroth-order features}.

High-order features in sequence prediction can be extended
from low-order ones by increasing the number of labels considered.
Formally for any label set $\Sigma$ and any feature vector $\bphi \in \mathbb{R}^d$,
the feature vector \term{lexicalized} with a label $s \in \Sigma$ is
defined as $\bphi \otimes \mathbbm{1}_s$,
where $\mathbbm{1}_s$ is a one-hot vector of length $|\Sigma|$
for the label $s$ and $\otimes: \mathbb{R}^{m \times n}
\times \mathbb{R}^{p \times q} \to \mathbb{R}^{mp \times nq}$
is the outer product.
With a slight abuse of notation, we let
$\bphi \otimes s = \bphi \otimes \mathbbm{1}_s$.
The resulting vector is
of length $|\Sigma|d$.  Similarly, we can lexicalize a feature vector
with pairs of labels, $\bphi \otimes s_1 \otimes s_2 = \bphi \otimes \mathbbm{1}_{s_1}
\otimes \mathbbm{1}_{s_2}$, giving a vector of length $|\Sigma|^2 d$.

For example, a common type of zeroth-order segmental feature
is of the form $\psi(x, t_1, t_2)$
where $x$ is the sequence of acoustic feature vectors, $t_1$
is the start time of the segment, and $t_2$ is the end time
of the segment.  To make it discriminative in a decoding graph $H$, we can compute the
first-order feature $\bphi_H(x, e)$ for any edge $e$
by first computing $\psi(x, t(\tail(e)), t(\head(e)))$ and then
lexicalizing it with the label $o_H(e)$.

To have a unified way of extending the order of features,
we define the concept of FST \term{structured composition},
or \term{$\sigma$-composition} for short, as follows.
For any two FSTs $A$ and $B$, the \term{$\sigma$-composed} FST is defined as
\begin{equation}
G = A \circ_\sigma B
\end{equation}
where
\begin{align}
V_G & = V_A \times V_B \\
E_G & = \bigg\{\langle e_1, e_2 \rangle \in E_A \times E_B:
    o_A(e_1) = i_B(e_2)\bigg\}
% \\
% \KGcomment{ } E_G & = \bigg\{e : e_1 \in E_A, e_2 \in E_B, 
%     o_A(e_1) = i_B(e_2), \nonumber\\
% &i_G(e) = i_A(e_1), o_G(e) = o_B(e_2), \nonumber\\
% &\tail_G(e)=(\tail_A(e_1),\tail_B(e_2)), \nonumber\\
% &\head_G(e)=(\head_A(e_1), \head_B(e_2))\bigg\}\nonumber
\end{align}
and
\begin{align}
i_G(\langle e_1, e_2 \rangle) & = i_A(e_1) \\
o_G(\langle e_1, e_2 \rangle) & = o_B(e_2) \\
\tail_G(\langle e_1, e_2 \rangle) & = \langle \tail_A(e_1), \tail_B(e_2) \rangle \\
\head_G(\langle e_1, e_2 \rangle) & = \langle \head_A(e_1), \head_B(e_2) \rangle
\end{align}
where $\langle \cdot, \cdot \rangle$ denotes a tuple.
\KGcomment{I'm a little uncomfortable with the 4 definitions above and Eq 12 being outside Eq 7; it makes it feel as if the output of sigma-composition must ``remember'' the internal labeling/structure of each edge even after defining the edge set. Do you need to make the internal structure of an edge persist forever, or can it be forgotten after $E_G$ and $w_G$ are defined? It would be nice if it could be forgotten, but based on the discussion below, it doesn't seem like it can be. So then is $G$ really a decoding graph, or a specialized decoding graph? Relatedly, I feel like the convention above of ``naming'' an edge in $E_G$ as $(e_1, e_2)$ is hard for the reader to wrap their head around (at least it's hard for me), because it suggests that $e_1$ is somehow the tail of the edge and $e_2$ is the head, but this is not the case since those types don't type-check (and as clarified in Eqs 10 and 11). It would be nice to use a different naming convention above if possible. I've attempted to do all this below Eq 7 above, but then there's the issue of defining $w_G$ outside the composition operation, which requires that we ``remember'' the edge pair corresponding to each new edge in $G$. So maybe changing the edge name above from $e$ to, e.g., $e_{1,2}$ or $e1|e2$ or $e_1^2$ might help.} 
% Comparing to
Unlike in classical composition, we only constrain the structure of $G$ and are free to define $w_G$ differently.
In particular, we let
\begin{equation}
w_G(\langle e_1, e_2 \rangle) = \btheta_G^\top \bphi_G(x, \langle e_1, e_2 \rangle),
\end{equation}
and $\bphi_G$ is free to use $\bphi_A$ and $\bphi_B$ but is
not constrained to do so.
In other words, the weight function $w_G$ can extract richer features than $w_A$
and $w_B$.

With structured composition, we can easily convert
low-order features to high-order ones.  Continuing
the above example, we can $\sigma$-compose the decoding
graph $H$ with a bigram language model (LM) $L$ in its FST form \cite{AllauzenEtAl2003}
with a slight modification.  We require the output labels of the LM FST to include the
history labels alongside the current label.
For example, the output labels of a bigram LM are of the form
$s_1s_2 \in \overline{\Sigma} \times \Sigma$,
where $s_1$ is the history label (possibly $\epsilon$) and $s_2$ is the current label.
Let $G = H \circ_\sigma L$.
% The feature $\bphi_G$ gets to access
% the frames $x$, and a pair of edges $\langle e_1, e_2 \rangle \in E_H \times E_L$.
We can define $t_G(\langle e_1, e_2 \rangle) = t_H(e_1)$.
For an edge $e \in E_G$,
we can compute first-order features $\varphi \otimes s_1$,
and second-order features $\varphi \otimes s_1 \otimes s_2$
for $s_1s_2 = o_G(e)$ and $s_1 \neq \epsilon$,
where $\varphi = \psi(x, t_G(\tail_G(e)), t_G(\head_G(e)))$.
If $s_1 = \epsilon$, everything falls back to the previous example.
% In sum, if we want to make use of all zeroth-, first-, and second-order
% features, we end up having
% \begin{equation}
% \bphi_G(x, (e_1, e_2)) = \begin{bmatrix}
%   \varphi \\
%   \varphi \otimes o_H(e_1) \\
%   \varphi \otimes o_H(e_1) \otimes \tail_L(e_2) \\
% \end{bmatrix}
% \end{equation}
% where $\varphi = \psi(x, t_H(\tail_H(e_1)), t_H(\head_H(e_1)))$. 
In general, by $\sigma$-composing
with high-order $n$-gram LMs, we can compute high-order features
by lexicalizing low-order ones. 
%\KLcomment{to me this still feels a bit like more notation/terminology than is needed to talk about low-order and high-order featues.  Also I wonder about referring to the head or tail of an edge, because there could be multiple FSAs with different edges that encode the same LM -- would you be using different feature functions if you sigma-compose with one than with another?}

\section{Discriminative Segmental Cascades}
\label{sec:approx}

Our approach, which we term a discriminative segmental cascade (DSC), is an instance of multi-pass decoding,
consisting of levels with increasing complexity of features and decreasing size of search space.  We start with the full search space and a ``simple'' first-level discriminative segmental model using inexpensive features, and use the first-level model to prune the search space.  We then apply a model using more expensive features, and optionally repeat the process for as many levels as desired.
%Inference with low-order features comes first,
%followed by inference with high-order features.
%We avoid the high computational cost of expensive features by pruning
%the search space before increasing the feature complexity.
%, which leads to the idea of a cascade.
Rather than the typical beam pruning, 
%To generate the cascade, we choose to prune the space with 
we prune with \term{max-marginals} \cite{SixtusOrtmanns1999,WeissEtAl2012}, which have certain useful properties and turn out to be important for achieving good performance with our models.
A max-marginal of an edge $e$ in $G$ is defined as
\begin{equation}
\gamma(e) = \max_{y \ni e}\; \btheta^\top \bphi(x, y).
\end{equation}
In words, it is the highest score of a path that passes through
the edge $e$.  We prune the edge if its max-marginal is
lower than a threshold, and keep it otherwise.  
%Since we would like 
In order to prune a multiplicative factor of edges at
each level of the cascade, Weiss et al. \cite{WeissEtAl2012} propose to use
the threshold
\begin{equation}
\tau_\lambda = (1 - \lambda) \frac{1}{|E_G|}\sum_{e \in E_G} \gamma(e)
    + \lambda \max_{y \in \mathcal{Y}} \btheta^\top \bphi(x, y),
\end{equation}
which interpolates between the mean of the max-marginals and
the maximum.
If $\lambda$ is set to 1, we only
keep the best path.

Lattice generation by max-marginal pruning guarantees that there is always at least one
path left after pruning and that any $y$ satisfying
$w(y) > \tau_\lambda$ is kept, because for every $e \in y$,
$\gamma(e) \ge w(y) > \tau_\lambda$.
In particular, if the ground truth has a score higher than
the threshold, it will still be in the search space
for the next level of the cascade.

Computing max-marginals in a specific level of the cascade
requires a forward pass and a backward pass through the graph.
Pruning with max-marginals thus takes twice the
time as searching for the best path alone. \KLcomment{check my rewording.}

Learning the cascade of models is also done level by level.
We start with the entire hypothesis space $H_1$ limited only by
a maximum segment length.  A first set of computationally inexpensive features
up to first order is used
for learning.  Let the first set of weights learned be
$\btheta_1$.  We can use $\btheta_1$ for first-pass decoding
if it is good enough, or we can choose to generate
the next level of the cascade and use more computationally
expensive features, such as higher-order ones.
Moving to the next level of the cascade,
we compute max-marginals with $\btheta_1$
and prune $H_1$ with a threshold, resulting in a lattice $H_2$.
If we wish increase the order of features, we $\sigma$-compose
$H_2$ with a bigram LM $L_2$.
A second set of features up to second order can then be used
for learning.  Suppose the second set of weights
is $\btheta_2$.  Again, we have the choice
either to use $\btheta_2$ for decoding
or to prune and repeat the process with more computationally expensive features.

\WWcomment{Moving to the next level does not necessarily mean you need to raise the order of features, correct? Maybe you simply want to reduce the hypothesis space and then use features that are more expensive to compute but of the same order?}
\HTcomment{Yes. The paragraph above is more about expensive features than about order.}

\vspace{-0.1in}
\section{EXPERIMENTS}
\label{sec:exp}

\KLcomment{say something about computation?  e.g. like the decoding times table in the MSLD poster?}

We experiment with segmental models in the context of phonetic recognition
on the TIMIT corpus~\cite{garofolo1993darpa}.
We follow the standard TIMIT protocol for training and
testing.  We use 192 randomly selected utterances
from the complete test set other than the core test set
as our development set,
and will refer to the core test set
simply as the test set.
The phone set is collapsed from 61 labels
to 48 before training.  In addition to the 48 phones,
we also keep the glottal stop /q/, sentence
start, and sentence end so that every frame in the training
set has a label. 
A summary of prior first-pass decoding results with segmental models,
along with our results and one from a standard speaker-independent HMM-DNN, is shown in
Table~\ref{tbl:summary}.  

\begin{table}
\caption{\label{tbl:summary} A summary of results in terms of phonetic error rate (\%)
    on the TIMIT test set, for prior first-pass segmental models, a speaker-independent HMM-DNN system given
    by a standard Kaldi recipe~\cite{povey2011kaldi}, and our models. }
\vspace{-0.1in}
\begin{center}
\begin{tabular}{lll}
              & dev      & test \\
              & PER (\%) & PER (\%)\\
\hline
HMM-DNN       &         & 21.4 \\
\hline
first first-pass SCRF \cite{Zweig2012}
              &         & 33.1 \\
Boundary-factored SCRF \cite{HeFoslerLussier2012}
              &         & 26.5 \\
Deep segmental NN \cite{AbdelHamidEtAl2013}
              &         & 21.87 \\
\hline
our first-pass model ($H_1$)
              & 22.15   & 21.73 \\
\hdashline
DSC 2$^{nd}$ level 
%($H_2 \circ_\sigma L_2$) 
with bigram LM
              & 19.80   &         \\
+ 2nd-order boundary features
              & 19.22   &         \\
+ 1st-order segment NN 
              & 18.86   &         \\
+ 1st-order bi-phone NN bottleneck
              & 18.77   & 19.93 \\
\end{tabular}
\end{center}
\vspace{-0.3in}
\end{table}

\subsection{First-pass segmental model}
\label{sec:firstpass}

\KLcomment{added some stuff about prior work}
First we demonstrate the effectiveness of our
first-pass decoder.
The first-pass search graph, denoted $H_1$,
contains all possible labels and all possible segmentations
up to 30 frames per segment.  Like some prior segmental phonetic recognition models~\cite{AbdelHamidEtAl2013,HeFoslerLussier2012}, many of the features in our first-pass decoder are based on averaging and sampling the outputs of a neural network phonetic frame classifier, specifically
a convolutional neural network (CNN) \cite{SimonyZisser2014}, which we describe next.

\subsubsection{CNN frame classifier}

The input to the network is a window of 15 frames of log-mel
filter outputs.
The network has five convolutional layers,
with 64--256 filters of size $5 \times 5$ for the input and $3 \times 3$ for others,
each of which
is followed by a rectified linear unit (ReLU) \cite{ZeilerEtAl2013} activation, with max pooling layers after
the first and the third ReLU layers.
The output of the final ReLU layer is concatenated with
a window of 15 frames of MFCCs centered on the current frame, and the resulting vector is passed
through three fully connected
ReLU layers with 4096 units each.
The network is trained with SGD for 35 epochs with a batch size
of 100 frames.  Fully connected layers and the concatenation layer
are trained with dropout at a 20\% and 50\% rate, respectively.
This
classifier was tuned on the development set and achieves a 22.1\% frame
error rate (after collapsing to 39 phone labels) on the test set.
We will use $\text{CNN}(x, t)$ to denote the log
of the final softmax layer, corresponding to the
predicted log probabilities of the phones, given as input
$[x_{t-7}; \dots; x_{t+7}]$.

\subsubsection{First-order features}

Below we list the features for each edge $(u, v)$.
We will use $L = t(v) - t(u)$ for short.

\begin{description}[leftmargin=0cm]
\item[Average of CNN log probabilities] 
    The log of the CNN output layer is averaged over all frames in the segment:
    \begin{equation}
    \frac{1}{L} \sum_{i=0}^{L-1} \text{CNN}(x, t(u) + i)
    \end{equation}

\item[Samples of CNN log probabilities]
    The log of the CNN output layer is sampled from
    the middle frames of three equally split sub-segments, i.e., 
    \begin{equation}
    \text{CNN}\left(x, t(u)
        + \left\lfloor \frac{[k + (k+1)]L}{3 \cdot 2} \right\rfloor \right)
    \end{equation}
    for $k = 0, 1, 2$.

\item[Boundary features]
    The log probabilities $i$ frames before the left boundary
    $\text{CNN}(x, t(u)-i)$ and $i$ frames after the right boundary
    $\text{CNN}(x, t(v)+i)$ are used as features.
    We use the concatenation of the boundary features for $i = 1, 2, 3$.

\item[Length indicator] 
%    We use indicator
    $\mathbbm{1}_{L = \ell}$
%    as our feature 
for $\ell = 0, 1, \dots, 30$.

\item[Bias]
%    The bias feature is just a
A constant 1.

\end{description}

We lexicalize all of the above features to first order, and include a
zeroth-order bias feature.
We minimize hinge loss with the overlap cost function introduced in \cite{TangEtAl2014}\WWcomment{We could mention overlap cost in section 2 $\cost(y,y')$} \KLcomment{yes}
with AdaGrad for up to 70 epochs
with step sizes tuned in $\{0.01, 0.1, 1\}$.
No explicit regularizer is used; instead we choose the step size and iteration that perform best on the
development set (so-called early stopping).
As shown in 
%the first row of
Table~\ref{tbl:summary}, our first-pass segmental model outperforms 
all previous segmental model TIMIT results of which we are aware.

\subsection{Higher-order features and segmental cascades}
\label{sec:highorder}

We next explore multi-pass decoding with beam search and with discriminative segmental cascades.  
In the second pass we include features of order two and a bigram LM $L_2$.
Back-off is approximated with $\epsilon$ transitions in the bigram LM.
Let $G = H \circ_\sigma L_2$,
where $H$ can be $H_1$ or $H_2$, the pruned $H_1$.
We consider the following additional features on edges $e \in E_G$.

\begin{description}[leftmargin=0cm]
\item[Bigram LM score]
    The bigram log probability
    $\log p_{\text{LM}}(s_2 | s_1)$,
%    is used as a feature, 
where $s_1s_2 = o_G(e)$
    We do not lexicalize this feature because it is naturally second-order.
\end{description}

\subsubsection{Beam search}

\begin{figure}
{\center
\includegraphics[width=1.6in]{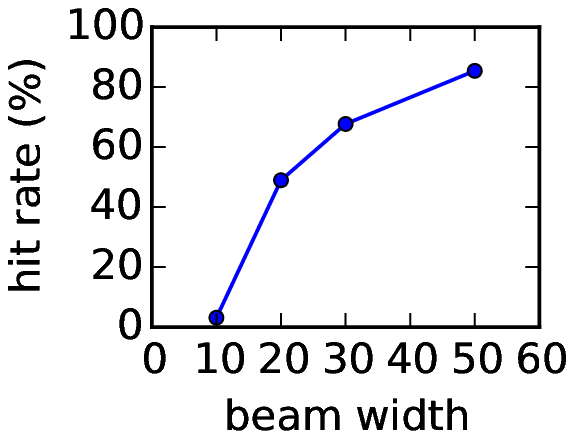}
\includegraphics[width=1.55in]{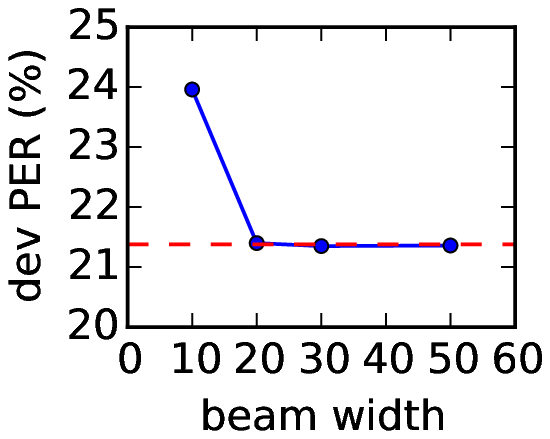}
\caption{\label{fig:beam-decode} Beam search on $H_1$ with
    different beam widths.  \emph{Left}: Hit rate
    on the development set.  \emph{Right}: PER on the development set.
    The dashed line is the PER of the exact search.}
}
\end{figure}

Before experimenting with the second-order features, we compare beam search
and exact search on the best model for $H_1$ to give a sense
of the approximation quality of beam search. 
%We run beam search with different beam widths.
We measure the quality of approximation
via the ``hit rate'', i.e., how often the exact best path is found.  Results are shown
in Figure~\ref{fig:beam-decode}.
As expected, the hit rate
decreases as the beam width decreases.
However, the PER does not decrease significantly, which
demonstrates that beam search is a good approximate decoding
algorithm when the model is well-trained. 

\begin{figure}
{\center
\includegraphics[width=1.65in]{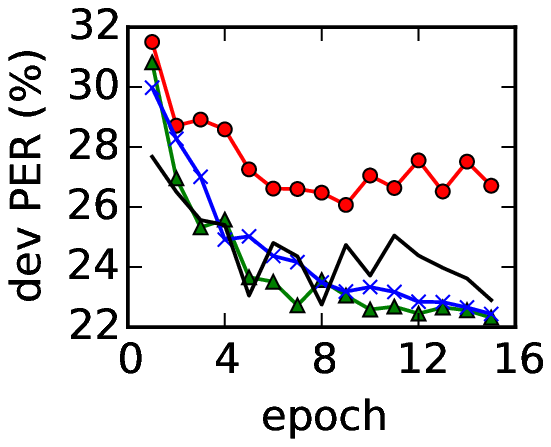}
\includegraphics[width=1.65in]{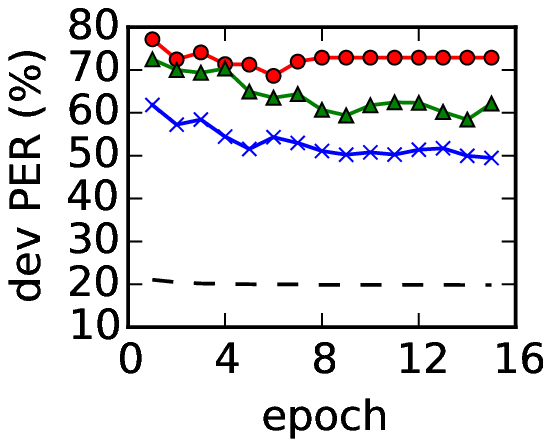}
\caption[]{\label{fig:beam-learn} Beam search for learning
    with different beam widths:
    \begin{tikzpicture}[baseline=-\the\dimexpr\fontdimen22\textfont2\relax]
    \node [right] at (0.2, 0) {beam=10};
    \draw[red] (-0.2, 0) -- (0, 0);
    \draw[red] (0, 0) -- (0.2, 0);
    \filldraw[fill=red,draw=black,radius=0.06] (0, 0) circle;
    \end{tikzpicture}
    \begin{tikzpicture}[baseline=-\the\dimexpr\fontdimen22\textfont2\relax]
    \node [right] at (0.2, 0) {beam=20};
    \draw[black!50!green] (-0.2, 0) -- (0, 0);
    \draw[black!50!green] (0, 0) -- (0.2, 0);
    \filldraw[fill=black!50!green,draw=black] (-0.07, -0.04) -- (0.07, -0.04)
        -- (0, 0.08) -- cycle;
    \end{tikzpicture}
    \begin{tikzpicture}[baseline=-\the\dimexpr\fontdimen22\textfont2\relax]
    \node [right] at (0.2, 0) {beam=30};
    \draw[blue] (-0.2, 0) -- (0, 0);
    \draw[blue] (0, 0) -- (0.2, 0);
    \draw[blue] (-0.07, 0.07) -- (0.07, -0.07);
    \draw[blue] (-0.07, -0.07) -- (0.07, 0.07);
    \end{tikzpicture}
    \begin{tikzpicture}[baseline=-\the\dimexpr\fontdimen22\textfont2\relax]
    \node [right] at (0.2, 0) {exact.};
    \draw (-0.2, 0) -- (0, 0);
    \draw (0, 0) -- (0.2, 0);
    \end{tikzpicture}
    \emph{Top}: Learning on $H_1$.
    \emph{Bottom}: Learning on $H_1 \circ_\sigma L_2$.  The
    dashed line is the learning curve of the second-level cascade
    $H_2 \circ_\sigma L_2$.
    \KLcomment{what are the oracle error rates with different k?}
    \HTcomment{I've never calculated.}}
}
\end{figure}

Judging from the decoding results, we use
beam search with beam widths $\{10, 20, 30\}$ for learning.
Since the runtime of beam search is controlled by the beam width
when the decoding graph is large, we can search directly on
$H_1 \circ_\sigma L_2$.  The composition is done on the fly
to avoid enumerating all edges in $H_1 \circ_\sigma L_2$.
We compare learning on both $H_1$
and $H_1 \circ_\sigma L_2$.  For $H_1$ we use the
same features as the first-pass segmental model, while for $H_1 \circ_\sigma L_2$
we add the bigram LM score and second-order boundary features. 
%\KLcomment{so here the $H_1$ features are not tied?} \HTcomment{Yes.}
For consistency, we use the same beam width for decoding.
Hinge loss is minimized with AdaGrad with 
step sizes tuned in \{0.01, 0.1, 1\}.
%We the one that achieves the lowest PER on the development set.
Results are shown in Figure~\ref{fig:beam-learn} for the step size that achieves the lowest development set PER.
When we train the segmental model on $H_1$
(top of Figure~\ref{fig:beam-learn}), learning
with beam search is successful when the beam width is large enough,
while for $H_1 \circ_\sigma L_2$ (bottom of Figure~\ref{fig:beam-learn}),
learning completely fails. \KLcomment{did you ever try larger beam widths for the larger search space? was it infeasible?} \HTcomment{I didn't try futher because it already takes 4 hrs per epoch for beam widths of 50.}

\subsubsection{Discriminative segmental cascades (DSC)}

\begin{figure}
{\center
\includegraphics[width=1.4in]{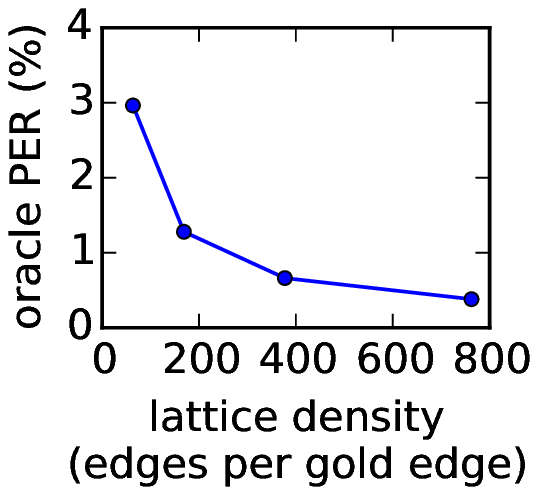}
\includegraphics[width=1.55in]{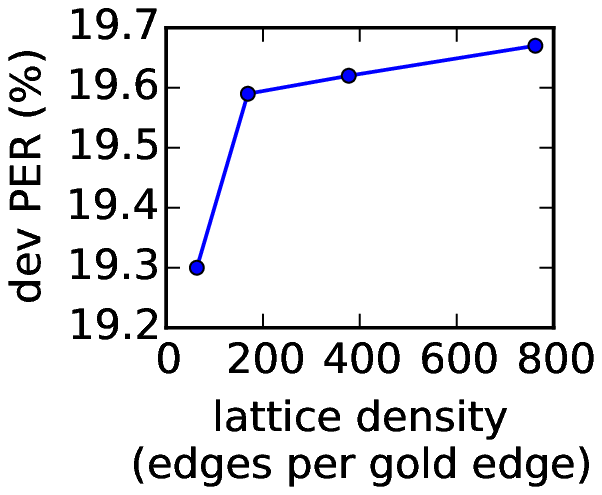}
\caption{\label{fig:density} Quality of $H_2$ for $\lambda$'s
    in \{0.8, 0.7, 0.6, 0.5\}.  
%The larger the $\lambda$, the     sparser the lattice.
    \emph{Left}: Oracle error rates for different lattice densities.\WWcomment{Have we defined oracle PER?} \KLcomment{I think it is OK to assume the readers know.}
    \emph{Right}: Corresponding second-pass development set PERs?}
}
\end{figure}

\KLcomment{reworded/shortened a bit here}
We next consider the proposed discriminative structured cascades (DSC) for utilizing the bigram LM and
second-order features.
We first prune $H_1$ with max-marginal pruning using our first-pass segmental model with weights $\btheta_1$,
resulting in $H_2$, and $\sigma$-compose $H_2$ 
with $L_2$.
%Several $\lambda$'s are tried for the pruning threshold.
Recall that the larger the pruning parameter $\lambda$, the sparser the lattice.
We measure the density of the lattice by the number of
edges in $H_2$ divided by the number of ground-truth (gold) edges.
The quality of $H_2$'s produced with different $\lambda$'s is
shown in Figure~\ref{fig:density} (left).
For the DSC second level, we define an additional feature:
\begin{description}[leftmargin=0cm]
\item[Lattice score]
    Instead of re-learning all of the weights for the features in the
    first-pass model, we combine them into 
    an additional feature from the first level of the cascade
    $\btheta_1^\top \bphi_{H_1}(x, e_1)$,
    which is never lexicalized, where $e_1 \in H$ is
    such that $\langle e_1, e_2 \rangle \in E_G$.
\end{description}
To compare with beam search, we use the lattice score, the bigram LM score,
second-order boundary features, first-order length indicators,
and first-order bias as our features for the second level of
the cascade. 
Hinge loss is minimized with AdaGrad
for up to 20 epochs with step sizes optimized in \{0.01, 0.1\}.
Again, no explicit regularizer is used except early stopping on
the development set.
Learning results on different lattices are shown in 
Figure~\ref{fig:density} (right).
We see that learning with the DSC is clearly better than with beam search.

\subsubsection{Other expensive features}

To add more context information,
we use the same CNN architecture and training setup to learn a bi-phone frame classifier,
but with an added 256-unit bottleneck linear layer
before the softmax \cite{SainathEtAl2013}.
Each frame is labeled with its segment label and one additional label from
a neighboring segment.  If the current frame is in the first half of the segment,
the additional label is the previous phone; if it is in the second half, 
then the additional label is the next phone.
%The network is trained in the same fashion.
The learned bottleneck layer outputs are used to define features (although they do not correspond to log probabilities) with averaging and sampling as for
the uni-phone case.  We refer to the resulting features as \term{bi-phone NN bottleneck} features.

Finally, we also use the same type of CNN
to train a {\emph segment} classifier.  Here the features at the input layer are
the log-mel
filter outputs from a 15-frame window around the segment's central frame.
The network architecture is the same as our frame classifier, but instead of
concatenation with 15-frame MFCCs, we concatenate with a segmental feature vector
%the segment features in \cite{ClarksonMoreno1999}
%because we know exactly where the segment starts and ends.
consisting of the average MFCCs of three sub-segments in
the ratio of 3-4-3, plus two four-frame averages
at both boundaries and length indicators for length 0 to 20 (similar to the segmental feature vectors of~\cite{Halberstadt1998,ClarksonMoreno1999}).
This CNN is trained on the ground-truth segments in the training set.
Finally, we build an ensemble of such networks
with different random seeds and a majority vote.
This ensemble classifier has a 15.0\% classification error on the test set,
which is to our knowledge the best result to date on the task of TIMIT phone segment classification (see Table~\ref{tbl:seg}). \KLcomment{we could remove the table and just cite the corresponding papers here}

\begin{table}
\caption{\label{tbl:seg} TIMIT segment classification error rates (ER).}
\begin{center}
\begin{tabular}{ll}
             & test ER (\%) \\
\hline
Gaussian mixture model (GMM) \cite{ClarksonMoreno1999}
             & 26.3 \\
SVM \cite{ClarksonMoreno1999}
             & 22.4 \\
Hierarchical GMM \cite{Halberstadt1998}
             & 21.0 \\
Discriminative hierarchical GMM \cite{ChangGlass2007}
             & 16.8 \\
SVM with deep scattering spectrum \cite{AndenMallat2014}
             & 15.9 \\
\hline
our CNN ensemble
             & 15.0
\end{tabular}
\end{center}
\vspace{-0.15in}
\end{table}

It is, however, still too time-consuming to compute the segment network outputs for every
edge in the lattice.  We instead compress the best-performing (single) CNN 
%from the ensemble \KLcomment{the ensemble?}
%\HTcomment{Weiran only took the best performing one I believe.}\WWcomment{Because the ensemble did not help.} 
into a shallow network with one hidden layer of 512 ReLUs
by training it to predict the log probability outputs
of the deep network, as proposed by~\cite{LiEtAl2014,BaCaruana2014}.
We then use the log probability outputs of the shallow network
and lexicalize them to first order.  We refer to the result as \term{segment NN} features.

Results with these additional features are shown in Table~\ref{tbl:summary}.
Adding the second-order features, bigram LM, and the above NN features
gives a 1.8\% absolute improvement over our best first-pass system, demonstrating the value of including such powerful but expensive features.

\vspace{-0.1in}
\section{Discussion}
\label{sec:disc}

We have presented discriminative segmental cascades (DSC), an approach for training and decoding with segmental models that allows us to incorporate high-order and complex features in a coarse-to-fine approach, and have applied them to the task of phone recognition.
The DSC approach uses max-marginal pruning, which outperforms beam search for learning the second-pass model.  Starting from a first-pass large-margin model that outperforms previous segmental model results and is competitive with HMM-DNNs, the DSC second pass improves the phone error rate by another 1.8\% absolute.

Further analysis may be needed to understand precisely why learning with beam search is not successful in the context of our models.
One issue is that $\sigma$-composing $H_1$ and $L_2$ introduces many
dead ends (paths that do not lead to final vertices) in the graph
because we have to do the composition on the fly.  Minimizing $H_1 \circ_\sigma L_2$
might help, but we would need to touch the edges of $H_1 \circ_\sigma L_2$ at least once, which is itself expensive. 
Second, even if we reach the final
vertices, the cost-augmented path might still have a lower
cost+score than the ground-truth path, which leads to
no gradient update.  This issue has been studied recently, and one possible solution is ``premature updates'' \cite{HuangEtAl2012}, but these are intended for the perceptron loss.
Third, the edge weights in our models are not strictly negative.
Beam search would tend to go depth-first when encountering
edges with positive weights.
On the other hand, if the edge weights are negative, beam search would tend
to go breadth-first,
which may explain why greedy search like beam search 
may cause problems for segmental models but works for HMMs.

Additional future work includes considering even more expressive features,
%We have seen that there is an obvious gain in using
%more context, such as
%second-order features and bi-phone frame classifiers.
%We can consider even 
higher-order features and additional cascade levels.
There is also much room for exploration with segment neural network classifiers.
One concern with our segment classifiers 
is that they are trained only with ground truth segments, so it is unclear how they behave when the input is
an incorrect hypothesized segment.  Alternatives include training on all hypothesized segments and allowing the network to learn to classify non-phones, similarly to the anti-phone and near-miss modeling of~\cite{Glass2003}.

\vspace{-0.1in}
\section{Acknowledgement}

This research was supported by NSF grant IIS-1433485. The opinions expressed in this work are those of the authors and do not necessarily reflect the views of the funding agency.

\cleardoublepage

\bibliographystyle{IEEEbib}
\bibliography{strings,refs}

\begin{thebibliography}{10}

\bibitem{ZweigNguyen2009}
Geoffrey Zweig and Patrick Nguyen,
\newblock ``A segmental {CRF} approach to large vocabulary continuous speech
  recognition,''
\newblock in {\em IEEE Workshop on Automatic Speech Recognition \&
  Understanding}, 2009, pp. 152--157.

\bibitem{SarawagiCohen2004}
Sunita Sarawagi and William~W Cohen,
\newblock ``Semi-{Markov} conditional random fields for information
  extraction,''
\newblock in {\em Advances in Neural Information Processing Systems}, 2004, pp.
  1185--1192.

\bibitem{ZhangGales2013}
Shi-Xiong Zhang and Mark Gales,
\newblock ``Structured {SVM}s for automatic speech recognition,''
\newblock {\em IEEE Transactions on Audio, Speech, and Language Processing},
  vol. 21, no. 3, pp. 544--555, 2013.

\bibitem{OstendorfEtAl1996}
Mari Ostendorf, Vassilios~V Digalakis, and Owen Kimball,
\newblock ``From {HMM}'s to segment models: {A} unified view of stochastic
  modeling for speech recognition,''
\newblock {\em IEEE Transactions on Speech and Audio Processing}, vol. 4, no.
  5, pp. 360--378, 1996.

\bibitem{Glass2003}
James Glass,
\newblock ``A probabilistic framework for segment-based speech recognition,''
\newblock {\em Computer Speech \& Language}, vol. 17, no. 2, pp. 137--152,
  2003.

\bibitem{TangEtAl2014}
Hao Tang, Kevin Gimpel, and Karen Livescu,
\newblock ``A comparison of training approaches for discriminative segmental
  models,''
\newblock in {\em Proceedings of the Annual Conference of International Speech
  Communication Association}, 2014.

\bibitem{ZweigEtAl2011}
Geoffrey Zweig, Patrick Nguyen, Dirk~Van Compernolle, Kris Demuynck, Les Atlas,
  Pascal Clark, Greg Sell, Meihong Wang, Fei Sha, Hynek Hermansky, Damianos
  Karakos, Aren Jansen, Samuel Thomas, G.S.V.S. Sivaram, Samuel Bowman, and
  Justine Kao,
\newblock ``Speech recognition with segmental conditional random fields: {A}
  summary of the {JHU} {CLSP} 2010 summer workshop,''
\newblock in {\em IEEE International Conference on Acoustics, Speech and Signal
  Processing}, 2011, pp. 5044--5047.

\bibitem{Zweig2012}
Geoffrey Zweig,
\newblock ``Classification and recognition with direct segment models,''
\newblock in {\em IEEE International Conference on Acoustics, Speech and Signal
  Processing}, 2012, pp. 4161--4164.

\bibitem{HeFoslerLussier2012}
Yanzhang He and Eric Fosler-Lussier,
\newblock ``Efficient segmental conditional random fields for phone
  recognition,''
\newblock in {\em Proceedings of the Annual Conference of the International
  Speech Communication Association}, 2012, pp. 1898--1901.

\bibitem{ZweigNguyen2010}
Geoffrey Zweig and Patrick Nguyen,
\newblock ``{SCARF}: {A} segmental conditional random field toolkit for speech
  recognition,''
\newblock in {\em Proceedings of the Annual Conference of International Speech
  Communication Association}, 2010, pp. 2858--2861.

\bibitem{AbdelHamidEtAl2013}
Ossama Abdel-Hamid, Li~Deng, Dong Yu, and Hui Jiang,
\newblock ``Deep segmental neural networks for speech recognition.,''
\newblock in {\em Proceedings of the Annual Conference of International Speech
  Communication Association}, 2013, pp. 1849--1853.

\bibitem{HeFoslerLussier2015}
Yanzhang He and Eric Fosler-Lussier,
\newblock ``Segmental conditional random fields with deep neural networks as
  acoustic models for first-pass word recognition,''
\newblock in {\em Proceedings of the Annual Conference of the International
  Speech Communication Association}, 2015.

\bibitem{OstendorfEtAl1991}
Mari Ostendorf, Ashvin Kannan, Steve Austin, Owen Kimball, Richard Schwartz,
  and Jan Rohlicek,
\newblock ``Integration of diverse recognition methodologies through
  reevaluation of n-best sentence hypotheses,''
\newblock in {\em Proceedings of the Workshop on Speech and Natural Language},
  1991, pp. 83--87.

\bibitem{WeissEtAl2012}
David Weiss, Benjamin Sapp, and Ben Taskar,
\newblock ``Structured prediction cascades,''
\newblock arXiv:1208.3279 [stat.ML], 2012.

\bibitem{AllauzenEtAl2003}
Cyril Allauzen, Mehryar Mohri, and Brian Roark,
\newblock ``Generalized algorithms for constructing statistical language
  models,''
\newblock in {\em Proceedings of the 41st Annual Meeting on Association for
  Computational Linguistics}, 2003, pp. 40--47.

\bibitem{SixtusOrtmanns1999}
A.~Sixtus and S.~Ortmanns,
\newblock ``High quality word graphs using forward-backward pruning,''
\newblock in {\em IEEE International Conference on Acoustics, Speech and Signal
  Processing}, 1999, pp. 593--596.

\bibitem{garofolo1993darpa}
John~S Garofolo, Lori~F Lamel, William~M Fisher, Jonathon~G Fiscus, and David~S
  Pallett,
\newblock ``Darpa timit acoustic-phonetic continous speech corpus cd-rom. nist
  speech disc 1-1.1,''
\newblock {\em NASA STI/Recon Technical Report N}, vol. 93, pp. 27403, 1993.

\bibitem{povey2011kaldi}
Daniel Povey, Arnab Ghoshal, Gilles Boulianne, Luk{\'a}{\v{s}} Burget,
  Ond{\v{r}}ej Glembek, Nagendra Goel, Mirko Hannemann, Petr
  Motl{\'\i}{\v{c}}ek, Yanmin Qian, Petr Schwarz, et~al.,
\newblock ``The {Kaldi} speech recognition toolkit,''
\newblock 2011.

\bibitem{SimonyZisser2014}
Karen Simonyan and Andrew Zisserman,
\newblock ``Very deep convolutional networks for large-scale image
  recognition,''
\newblock arXiv:1409.1556 [cs.CV], 2014.

\bibitem{ZeilerEtAl2013}
Matthew Zeiler, Marc'Aurelio Ranzato, Rajat Monga, Min Mao, Kun Yang, Quoc Le,
  Patrick Nguyen, Alan Senior, Vincent Vanhoucke, Jeffrey Dean, and Geoff
  Hinton,
\newblock ``On rectified linear units for speech processing,''
\newblock in {\em IEEE International Conference on Acoustics, Speech and Signal
  Processing}, 2013, pp. 3517--3521.

\bibitem{SainathEtAl2013}
Tara Sainath, Brian Kingsbury, Vikas Sindhwani, Ebru Arisoy, and Bhuvana
  Ramabhadran,
\newblock ``Low-rank matrix factorization for deep neural network training with
  high-dimensional output targets,''
\newblock in {\em IEEE International Conference on Acoustics, Speech and Signal
  Processing}, 2013, pp. 6655--6659.

\bibitem{Halberstadt1998}
Andrew Halberstadt,
\newblock {\em Heterogeneous Acoustic Measurements and Multiple Classifiers for
  Speech Recognition},
\newblock Ph.D. thesis, Massachusetts Institute of Technology, 1998.

\bibitem{ClarksonMoreno1999}
Philip Clarkson and Pedro Moreno,
\newblock ``On the use of support vector machines for phonetic
  classification,''
\newblock in {\em IEEE International Conference on Acoustics, Speech, and
  Signal Processing}, 1999, vol.~2, pp. 585--588.

\bibitem{ChangGlass2007}
Hung-An Chang and James Glass,
\newblock ``Hierarchical large-margin {Gaussian} mixture models for phonetic
  classification,''
\newblock in {\em IEEE Workshop on Automatic Speech Recognition \&
  Understanding}, 2007, pp. 272--277.

\bibitem{AndenMallat2014}
Joakim And{\'e}n and St{\'e}phane Mallat,
\newblock ``Deep scattering spectrum,''
\newblock {\em IEEE Transactions on Signal Processing}, vol. 62, no. 16, pp.
  4114--4128, 2014.

\bibitem{LiEtAl2014}
Jinyu Li, Rui Zhao, Jui-Ting Huang, and Yifan Gong,
\newblock ``Learning small-size {DNN} with output-distribution-based
  criteria,''
\newblock in {\em Proceedings of the Annual Conference of the International
  Speech Communication Association}, 2014.

\bibitem{BaCaruana2014}
Jimmy Ba and Rich Caruana,
\newblock ``Do deep nets really need to be deep?,''
\newblock in {\em Advances in Neural Information Processing Systems}, 2014, pp.
  2654--2662.

\bibitem{HuangEtAl2012}
Liang Huang, Suphan Fayong, and Yang Guo,
\newblock ``Structured perceptron with inexact search,''
\newblock in {\em Proceedings of the Conference of the North American Chapter
  of the Association for Computational Linguistics: Human Language
  Technologies}, 2012, pp. 142--151.

\end{thebibliography}

\end{document}